%% file: main.tex
\documentclass[10pt,twocolumn,letterpaper]{article}

\usepackage[pagenumbers]{cvpr} 

\input{preamble}

\title{MultiShotMaster: A Controllable Multi-Shot Video Generation Framework}

\author{
Qinghe Wang\textsuperscript{1$\dagger$} \quad
Xiaoyu Shi\textsuperscript{2}$^\text{\Letter}$ \quad
Baolu Li\textsuperscript{1} \quad
Weikang Bian\textsuperscript{3} \quad 
Quande Liu\textsuperscript{2} \quad 
Huchuan Lu\textsuperscript{1} \quad \\
Xintao Wang\textsuperscript{2} \quad  
Pengfei Wan\textsuperscript{2} \quad 
Kun Gai\textsuperscript{2} \quad
Xu Jia\textsuperscript{1}$^\text{\Letter}$\\[4pt]
\normalsize$^{1}$Dalian University of Technology~~
\normalsize$^{2}$Kling Team, Kuaishou Technology~~
\normalsize$^{3}$The Chinese University of Hong Kong\\[4pt]
\href{https://qinghew.github.io/MultiShotMaster}{\textcolor{magenta}{https://qinghew.github.io/MultiShotMaster}}
}

\begin{document}

\twocolumn[{
\renewcommand\twocolumn[1][]{#1}
\maketitle  
\vspace{-0.65cm}
\begin{center}
\centering
    \captionsetup{type=figure}
    \includegraphics[width=1\linewidth]{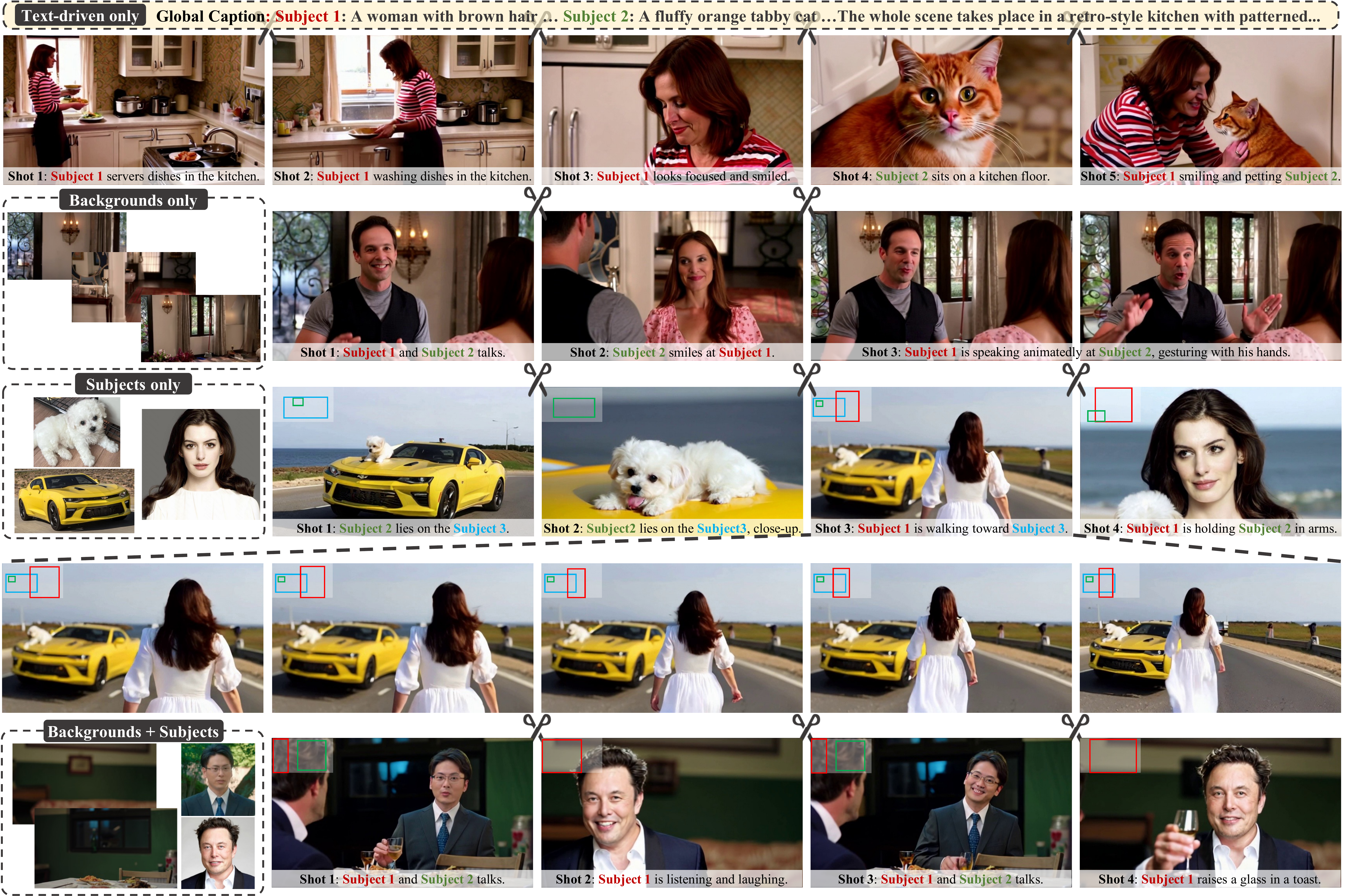}
    \vspace{-0.45cm}
    \captionof{figure}{We propose MultiShotMaster, the first controllable multi-shot video generation framework that supports text-driven inter-shot consistency, customized subject with motion control, and background-driven customized scene. Both shot counts and shot durations are variable. 
    Only the global caption of the first case is shown for brevity.
    }
    \label{fig:first_image}
\end{center}
}]

\let\thefootnote\relax\footnotetext{$\dagger$ Work done during an internship at Kling Team, Kuaishou Tech.} \let\thefootnote\relax\footnotetext{$\text{\Letter}$ ~Corresponding authors.}

\input{0_abstract}    
\input{1_intro}

\input{2_related_works}

\input{3_method}

\input{4_experiment}
\input{5_conclusion}

\input{main.bbl}

%

\clearpage

\appendix

\input{supp}

\end{document}

%% file: preamble.tex
\usepackage{makecell}
\usepackage{xcolor}
\usepackage{colortbl}
\definecolor{cvprblue}{rgb}{0.21,0.49,0.74}
\usepackage[pagebackref,breaklinks,colorlinks,allcolors=cvprblue]{hyperref}
\usepackage{graphicx}
\usepackage{array}
\usepackage{svg}

\definecolor{myorange}{RGB}{255,100,3}
\definecolor{mygray}{gray}{.85}
\definecolor{mygray1}{gray}{.7}
\definecolor{mygray2}{gray}{.93}
\definecolor{mygray3}{gray}{.90}
\newcolumntype{C}[1]{>{\centering\arraybackslash}p{#1}}
\usepackage{multirow}
\usepackage{multicol}
\usepackage{pifont}
\newcommand{\xmarkg}{\textcolor{gray}{\ding{55}}\xspace}%

\usepackage[misc]{ifsym} 

\usepackage{algorithm}
\usepackage{algorithmic}



%% file: 0_abstract.tex
\begin{abstract}
Current video generation techniques excel at single-shot clips but struggle to produce narrative multi-shot videos, which require flexible shot arrangement, coherent narrative, and controllability beyond text prompts. To tackle these challenges, we propose MultiShotMaster, a framework for highly controllable multi-shot video generation. We extend a pretrained single-shot model by integrating two novel variants of RoPE. First, we introduce Multi-Shot Narrative RoPE, which applies explicit phase shift at shot transitions, enabling flexible shot arrangement while preserving the temporal narrative order. Second, we design Spatiotemporal Position-Aware RoPE to incorporate reference tokens and grounding signals, enabling spatiotemporal-grounded reference injection. In addition, to overcome data scarcity, we establish an automated data annotation pipeline to extract multi-shot videos, captions, cross-shot grounding signals and reference images. Our framework leverages the intrinsic architectural properties to support multi-shot video generation, featuring text-driven inter-shot consistency, customized subject with motion control, and background-driven customized scene. Both shot count and duration are flexibly configurable. Extensive experiments demonstrate the superior performance and outstanding controllability of our framework.
\end{abstract}

%% file: 1_intro.tex
\section{Introduction}
\label{sec:intro}
Recently, riding the wave of success in video generation~\cite{videoworldsimulators2024}, content creators have produced many interesting videos and attracted massive traffic. Powered by diffusion transformers~(DiTs)~\cite{peebles2023scalable,wan2025wan,opensora,lin2024open,yang2024cogvideox}, the semantic information encapsulated in text prompts can be presented in a single-shot high-quality generated video. Furthermore, advances in controllability~\cite{ma2025controllable,zhang2025generative,ye2025unic} have endowed video generation with increased versatility and creative capabilities by incorporating diverse condition signals such as reference image~\cite{liu2025phantom,vace} and object motion~\cite{geng2025motion,wang2025cinemaster}. However, real-world films and television series consist of multi-shot video clips with narrative structures that convey coherent stories to audiences. Such storytelling relies on cinematic language, encompassing holistic scenes, character interactions, and microexpressions. Therefore, there is a significant gap between current video generation techniques and practical video content creation.

Ideally, the AI-powered video generation system should deliver director-level controllable multi-shot functionality to users, encompassing four critical aspects: (1) variable shot counts and flexible shot durations, (2) dedicated text descriptions for each shot, (3) character appearance and scene definition, (4) character movement control. The development of this vision is still in its infancy. Current multi-shot video generation methods typically fall into two paradigms: text-to-keyframe generation followed by image-to-video~(I2V) generation~\cite{zhou2024storydiffusion,xiao2025captain,yin2023nuwa,chen2023seine}, and direct end-to-end generation~\cite{wu2025cinetrans,kara2025shotadapter,guo2025long,qi2025mask,cai2025mixture}. The keyframe-based paradigm first generates a set of keyframes with visual consistency and then uses the I2V model to generate each shot with the corresponding keyframe. Although it has decent applicability, the limited conditional capability of sparse keyframes cannot cover the briefly-appearing characters and scene consistency outside the keyframes. The end-to-end paradigm can generate multi-shot videos directly, and keep the consistency by using full attention along the temporal dimension. Yet, it is still constrained by the fixed shot duration or limited shot counts. Moreover, both paradigms can only be driven by text prompts. As a result, there is an urgent need to investigate comprehensive controllability in multi-shot video generation, covering flexible shot arrangements and more condition signals.

Under the DiT architecture, all image patches are projected into token embeddings and concatenated together, typically requiring positional encoding (\eg, Rotary Position Embedding~(RoPE)~\cite{su2024roformer}) to preserve their spatiotemporal order. The RoPE-based attention mechanism has a crucial property: tokens with closer spatiotemporal distance receive higher attention weights, enabling the model to capture local spatiotemporal correlations. This property inspires our two insights: (1) using continuous RoPE to all frames of multi-shot videos in temporal order will cause the model to confuse intra-shot consecutive frames with inter-shot frames across shot boundaries. (2) applying the RoPE of a specified region to reference features will bridge the connection with the corresponding video tokens.

In this work, we present MultiShotMaster, the first framework for highly controllable multi-shot video generation. Specifically, we extend a pretrained single-shot text-to-video model to controllable multi-shot generation primarily through improvements in RoPE. First, we introduce Multi-Shot Narrative RoPE that breaks RoPE's continuity at shot transitions and helps the model recognize shot boundaries, enabling controllable shot transitions. Furthermore, we design Spatiotemporal Position-Aware RoPE, which incorporates spatiotemporal-grounded control signals into RoPE for reference tokens (subjects and backgrounds). It establishes strong correlations between reference tokens and the corresponding video tokens during attention, allowing reference injection into specified spatiotemporal regions. By specifying multiple frames, users can further control subject motion. To manage the in-context information flows between reference and video tokens, we develop a Multi-Shot \& Multi-Reference Attention Mask. In addition, we propose an automatic Multi-Shot \& Multi-Reference Data Curation pipeline to provide essential training data. Comprehensive evaluations demonstrate its superior performance and outstanding controllability. We integrate diverse conditions, including text prompts, subjects, grounding signals, and backgrounds for multi-shot video generation with flexible shot arrangement, as shown in Fig.~\ref{fig:first_image}. We anticipate that this work could inspire future research in controllable multi-shot video generation.

%% file: 2_related_works.tex
\begin{figure*}[!t]
  \centering
  \includegraphics[width=\linewidth]{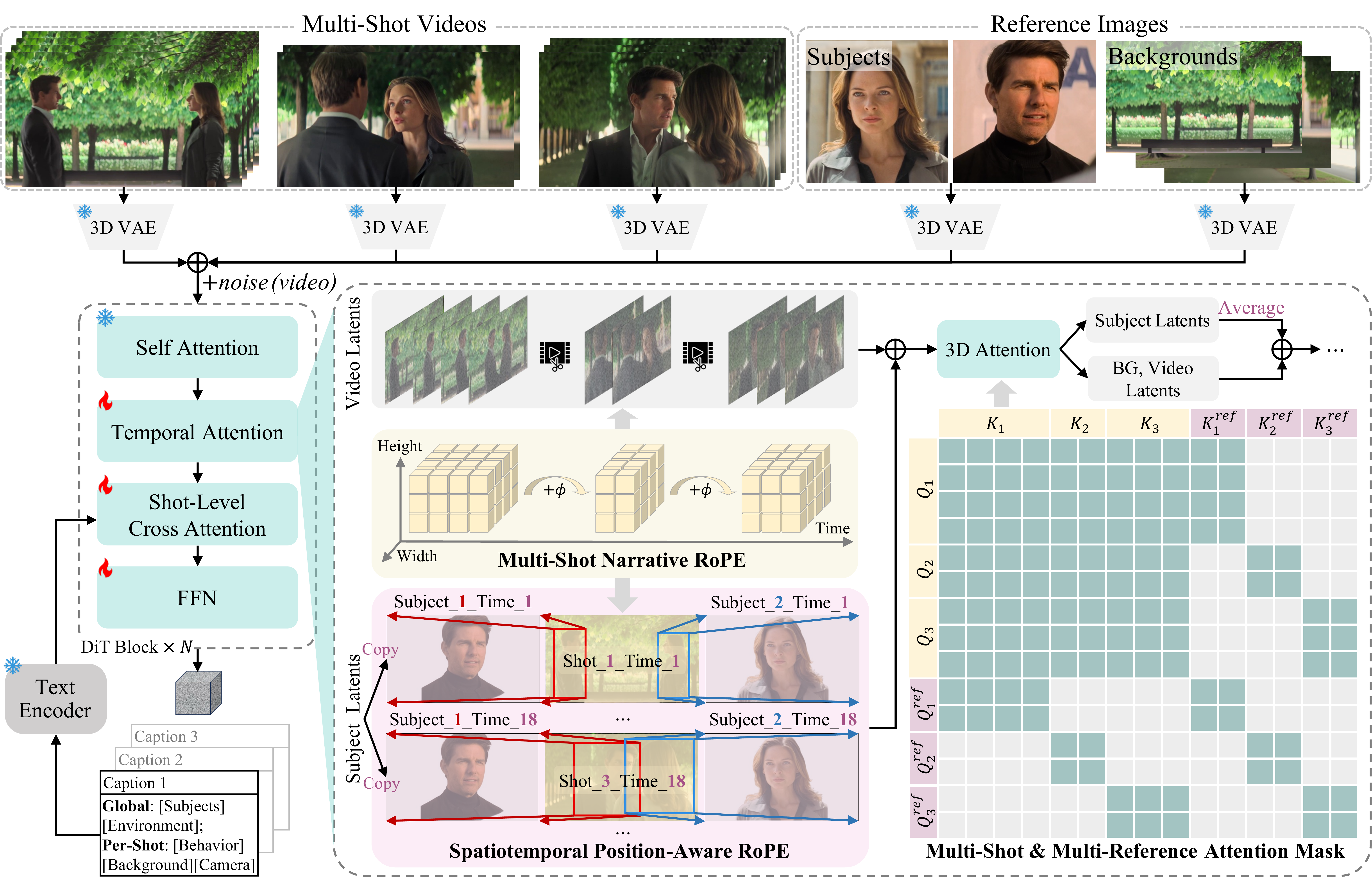}
  \vspace{-14pt} 
  \caption{\textbf{Overview of MultiShotMaster.} We extend a pretrained single-shot T2V model by two key RoPE variants: Multi-Shot Narrative RoPE for flexible shot arrangement with temporal narrative order, and Spatiotemporal Position-Aware RoPE for grounded reference injection. To manage in-context information flows, we design a Multi-Shot \& Multi-Reference Attention Mask. We finetune temporal attention, cross attention and FFN, leveraging the intrinsic architectural properties to achieve flexible and controllable multi-shot video generation.}
  \label{fig:framework}
  \vspace{-10pt} 
\end{figure*}

\section{Related Works}
\label{sec:related_works}
\subsection{Text-to-Video Generation}
Early methods inflated the pretrained text-to-image generation models~\cite{rombach2022high} with temporal layers~\cite{wu2023tune,animatediff} for video generation, achieving preliminary short video animation. Recent approaches have employed the diffusion transformer~(DiT)~\cite{peebles2023scalable} architectures, which could generate longer, high-quality videos with detailed text description~\cite{videoworldsimulators2024,gao2025seedance,wan2025wan,opensora,lin2024open,yang2024cogvideox}. However, scaling video generation beyond short clips remains an open problem. Existing research focuses on two distinct tasks: single-shot and multi-shot long video generation. The former typically encounters issues of error accumulation and memory loss~\cite{chen2024diffusion,huang2025self,zhang2025packing,huang2025memory,yu2025context}, while the latter focuses on narrative coherence and inter-shot consistency~\cite{xiao2025captain,qi2025mask,cai2025mixture,guo2025long}.

\subsection{Multi-Shot Video Generation}
Multi-shot videos should preserve narrative logic and ensure spatiotemporal consistency of character positioning and scene layout across all shots~\cite{he2025cut2next,meng2025holocine}. The existing paradigms mainly comprise text-to-keyframe generation \& image-to-video generation~\cite{zhou2024storydiffusion,xiao2025captain,zhang2025shouldershot} and end-to-end holistic generation~\cite{wu2025cinetrans,kara2025shotadapter,guo2025long,qi2025mask,cai2025mixture,wei2025mocha}. The former depends on the generation quality of keyframes and cannot cover the character and scene consistency outside the keyframes. The latter exhibits better consistency, benefiting from the full attention along the temporal dimension. We observe that the multi-shot videos with fixed shot duration might contain ``boring" frames. For instance, when a director wants to capture a close-up of an actor picking up a drink, such an insert shot requires only a few frames for audiences to understand the content. To tackle this problem, ShotAdapter~\cite{kara2025shotadapter} incorporates learnable transition tokens that interact only with shot-boundary frames to indicate transitions. CineTrans~\cite{wu2025cinetrans} constructs an attention mask to weaken the inter-shot correlations, enabling transitions at predefined positions. In contrast, our work conveys the transition signals by manipulating the RoPE embeddings, which prevents interference with token interactions in pretrained attention and explicitly achieves shot transitions.

\subsection{Controllable Video Generation}
Providing explicit and precise user control is essential for practical content creation~\cite{wang2025stableidentity, wang2025characterfactory, wu2024motionbooth, wei2024DreamVideo2}. The controllable video generation field supports diverse control signal types such as camera motion~\cite{bian2025gs,bai2025recammaster,bai2024syncammaster,he2024cameractrl}, object motion~\cite{shi2024motion,xing2025motioncanvas,ma2024follow,ma2025follow}, reference video~\cite{bian2025relightmaster,bian2025video,luo2025camclonemaster,li2025vfxmaster}. VACE~\cite{vace} and Phantom~\cite{liu2025phantom} support multi-reference video generation and achieve realistic composition. Tora~\cite{zhang2025tora} and Motion Prompting~\cite{geng2025motion} control the object motion through point trajectories. However, existing methods typically focus on the single-shot setting and adopt separate adapters for reference injection and motion control. If following the traditional paradigm~\cite{liu2025phantom,wang2025cinemaster}, controllability in multi-shot settings would require larger networks and incur higher computational costs. To address this limitation, we propose the first controllable multi-shot framework that supports reference injection and motion control jointly, requiring no additional adapters.

%% file: 3_method.tex
\section{Method}
\label{sec:method}
\subsection{Evolving from Single-Shot to Multi-Shot T2V}
\label{sec:3.1}
\textbf{Preliminary}: Our model is developed upon a pretrained single-shot text-to-video~(T2V) model with $\sim$1B parameters, which consists of a 3D Variational Auto-Encoder (VAE)~\cite{kingma2013auto}, T5 text encoder~\cite{raffel2020exploring} and a latent diffusion transformer~(DiT) model~\cite{peebles2023scalable}. Each basic DiT block contains a sequence including a 2D spatial self-attention module, a 3D spatiotemporal self-attention module, a text cross-attention module and a feed-forward network~(FFN). We define a straight path from clean data $z_0$ to noised data $z_\tau$ at timestep $\tau$ using Rectified Flow~\cite{esser2024scaling}: $z_t = (1-\tau)z_0 + \tau\epsilon$, where $\epsilon\in\mathcal{N}(0,\mathbf{I})$. The denoising process follows the ordinary differential equation: $dz_\tau = v_{\Theta}(z_\tau,\tau,c_{text})d\tau$, where $v_\Theta$ is the denoising network. The training objective is to regress velocity~\cite{lipman2022flow}:
\begin{equation}
\label{lcm_loss}
     \mathcal{L}_{LCM} = \mathbb{E}_{\tau,\epsilon,z_0}\left[\|(z_1-z_0)-v_{\Theta}(z_\tau,\tau,c_{text})\|^2_2\right]
\end{equation}

As shown in Fig.~\ref{fig:framework}, to adapt the input from single-shot to multi-shot videos with the sudden content changes at shot boundaries, we encode each shot separately through 3D VAE and then concatenate the video latents. During temporal attention, the original 3D-RoPE assigns sequential indices along the temporal dimension, leading to a critical issue: the model cannot distinguish between intra-shot consecutive frames with inter-shot frames across shot boundaries. To explicitly help the model perceive shot boundaries, we propose a \textbf{Multi-Shot Narrative RoPE} mechanism that introduces an angular phase shift into the original 3D-RoPE for each transition. The query~(Q) of $i$-th shot is computed as follows, and similarly for key~(K):
\begin{equation}
\label{multishotrope}
Q_i = \text{RoPE}((t + i\phi)\cdot f,\ h\cdot f,\ w\cdot f) \odot \tilde{Q}_i
\end{equation}
where $(t, h, w)$ are spatiotemporal position indices, $\phi$ is the angular phase shift factor, $f$ is the decreasing base frequency vector, and $\odot$ denotes the element-wise rotary transformation of query embeddings $\tilde{Q}_i$ via complex rotations. This design not only maintains the narrative shooting order of inter-shot frames, but also leverages RoPE's inherent rotational properties to mark shot boundaries through fixed phase shifts, requiring no additional trainable parameters. It enables users to flexibly configure both the number of shots and their respective durations.

Considering that providing users with the capability to customize each shot individually could facilitate content creation, we design a hierarchical prompt structure. It includes a global caption that describes subject appearances and environments, and per-shot captions that detail subject actions, backgrounds, and camera, following~\cite{guo2025long}. For each shot, we combine the global caption with the corresponding per-shot caption. In the vanilla T2V model, T5 encoder encodes the text prompts as text embeddings which are then replicated along the temporal dimension to align with the video frame sequence for text-frame cross attention. Accordingly, we replicate each shot's text embeddings to align with the corresponding shot frame count, enabling shot-level cross-attention.

\begin{figure*}[!t]
  \centering
  \includegraphics[width=\linewidth]{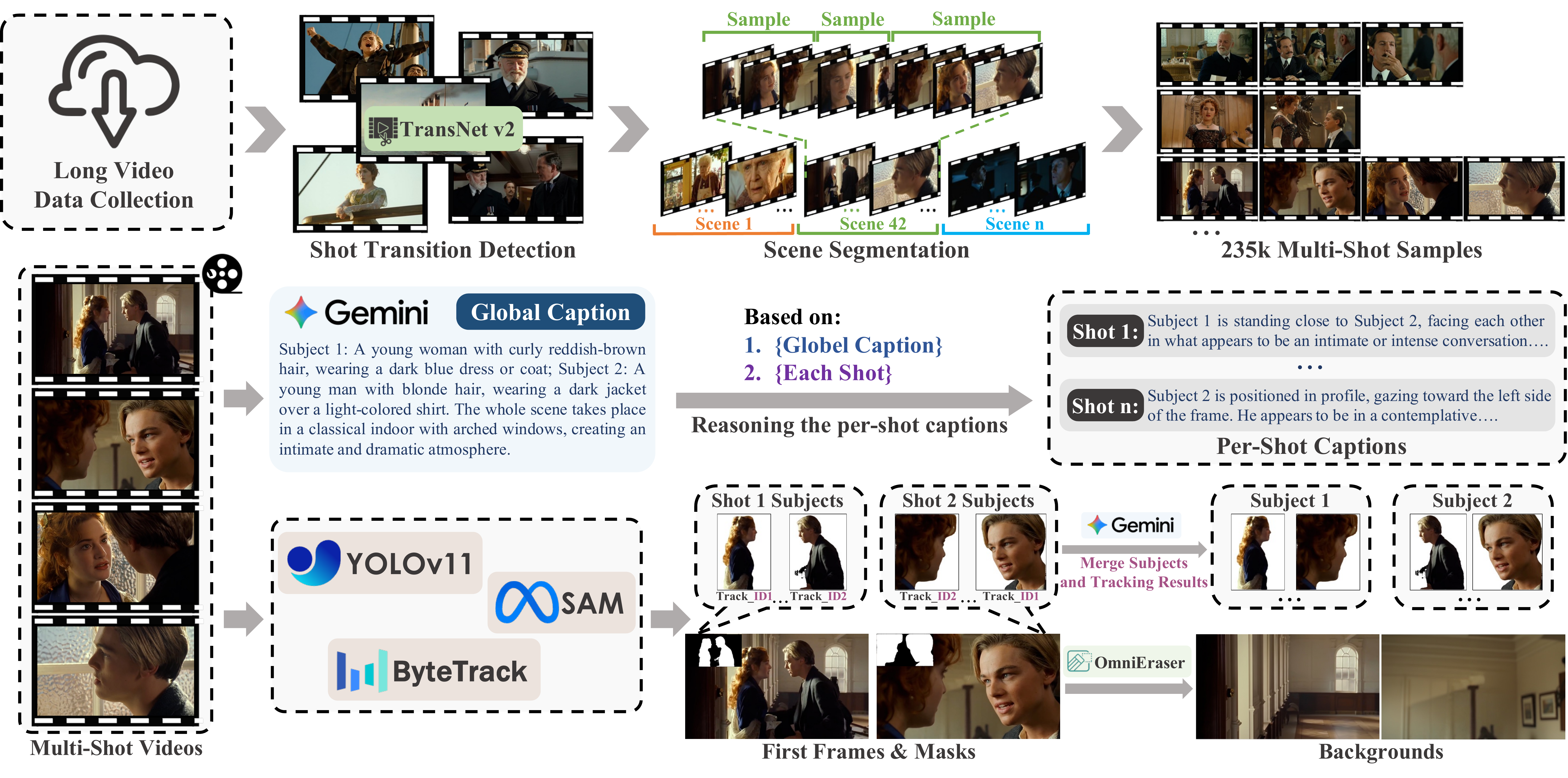}
  \vspace{-18pt} 
  \caption{\textbf{Data Curation Pipeline:} (1) We employ a shot transition detection model~\cite{soucek2024transnet} to cut the collected long videos into short clips, use a scene segmentation model~\cite{wu2022scene} to cluster clips within the same scene, and then sample multi-shot videos. (2) We introduce a hierarchical caption structure and use Gemini-2.5~\cite{comanici2025gemini} in a two-stage process to produce global caption and per-shot captions. (3) We integrate YOLOv11 \cite{khanam2024yolov11}, ByteTrack~\cite{zhang2022bytetrack} and SAM~\cite{kirillov2023segment} to detect, track and segment subject images. Then we use Gemini-2.5 to merge the per-shot tracking results by subject appearance. We obtain clean backgrounds by using OmniEraser~\cite{wei2025omnieraser}.
  } 
  \label{fig:data_pipe}
  \vspace{-10pt} 
\end{figure*}

\subsection{Spatiotemporal-Grounded Reference Injection}
\label{sec:3.2}
Users typically require the capability to provide reference images (e.g., subjects and backgrounds) and motion control signal for creating customized video content. To address this requirement, we propose \textbf{Spatiotemporal Position-Aware RoPE} to improve in-context learning for spatiotemporal-grounded reference injection. Specifically, we individually encode each reference image into the latent space through 3D VAE and concatenate them with the noised video latents via token concatenation. In temporal attention, clean reference tokens propagate visual information to noisy video tokens for reference injection. Furthermore, 3D-RoPE enables tokens with closer spatiotemporal distance to attend more to each other. Inspired by this mechanism, we apply 3D-RoPE from specified regions to the corresponding reference tokens, thereby establishing strong correlations between region-specified video tokens and reference tokens, as shown in Fig.~\ref{fig:framework}. Since the subject bounding box region~$(x_1,y_1,x_2,y_2)$ at $t$-th frame is smaller than the spatial dimensions $(H,W)$ of reference tokens, we sample the 3D-RoPE by:
\label{multiid_rope}
\begin{align}
Q^{ref} =& \ \text{RoPE}((t + i\phi)\cdot f,\ h^{ref}\cdot f,\ w^{ref}\cdot f) \odot \tilde{Q}^{ref}, \notag\\ 
h^{ref} &= y_1 + \frac{(y_2 - y_1)}{H} \cdot j, \quad j \in [0, H-1], \\
w^{ref} &= x_1 + \frac{(x_2 - x_1)}{W} \cdot k, \quad k \in [0, W-1] \notag
\end{align}
In this manner, we can control the subject to appear at a specified spatiotemporal position. To control the motion trajectory of a subject, we create multiple copies of the subject tokens and assign different spatiotemporal RoPE to each copy. The temporal attention then transfers the subject motion embedded in these copies to video tokens at the corresponding spatiotemporal positions. The copied tokens of each subject will be averaged after attention. Similarly, to achieve multi-shot scene customization, we copy the 3D-RoPE from the first frame of each shot and apply it to the corresponding background tokens. \textit{We further clarify the details in temporal attention in Appendix~\ref{supp:attnt}.}

By incorporating spatiotemporal-controllable multi-reference injection, we significantly expand the functional boundaries of multi-shot video generation, providing users with more flexible and powerful video creation capabilities. It allows users to customize characters using subject images and control their positions and movements, which improves the practicality of multi-shot video generation. In addition, using multiple background images of a scene can achieve customized multi-shot scene consistency, reducing the necessity for location shooting.

\subsection{Multi-Shot \& Multi-Reference Attention Mask}
\label{sec:3.3}
The user-provided reference images and subject copies may lead to excessively long contexts with high computational costs. On the other hand, there are unnecessary interactions between in-context tokens. For example, Subject 0 may only appear in shot 2, but unconstrained token concatenation allows other shots to access Subject 1 as well. Although 3D-RoPE can guide spatiotemporal-specified video tokens to attend to reference tokens, the small but non-zero attention weights still pose a content leakage risk. Therefore, we design a multi-shot \& multi-reference attention mask to constrain information flow and optimize attention allocation. As shown in Fig.~\ref{fig:framework}, we maintain full attention across all multi-shot video tokens for global consistency, and limit each shot to access only the reference tokens within its own shot. Accordingly, the reference tokens of each shot can only attend to each other and the video tokens within the same shot. This attention mask strategy effectively ensures that each shot focuses on the intra-shot reference injection, and keeps global consistency by inter-shot full attention.

\subsection{Training and Inference Paradigm}
\label{sec:3.4}
Typically, controllable capabilities are trained based on a foundation generation model. Considering that the reference injection task requires training on large-scale data to learn diverse subjects, and the construction cost of multi-shot \& multi-reference data is relatively high, we first train spatiotemporal-specified reference injection on 300k single-shot data. We sample bounding boxes with random starting points and 1-second intervals, where each bounding box has a 0.5 drop probability. This sparse bounding box sequence allows users to control subjects easily. In the second stage, we train on the constructed multi-shot \& multi-reference data. To enable controllable multi-shot video generation with text-, subject-, background-, and joint-driven modes, we randomly drop the subject and background, each with a 0.5 probability. We notice that the training objective  Eq.~\ref{lcm_loss} focuses on global consistency and ignores details. Therefore, we propose a cross-shot subject-focused post-training that assigns $(2\times)$ loss weight to subject regions versus $(1\times)$ to backgrounds. It not only improves subject consistency but also enables the model to better comprehend how the subjects change across different shots. \textit{More details could be found in Appendix~\ref{supp:train_paradigms}.}

During inference, our framework supports controllable multi-shot video generation with text-driven inter-shot consistency, customized subject with motion control, and background-driven customized scene consistency. Both shot count and duration are flexibly configurable. This versatile framework opens new possibilities for diverse multi-shot video content creation, enabling users to craft highly customized video narratives.

\subsection{Multi-Shot \& Multi-Reference Data Curation} 
\noindent\textbf{Multi-Shot Videos}: To construct a multi-shot video dataset, as shown in Fig.~\ref{fig:data_pipe}, we first crawl long videos from the Internet~(including diverse types: movies, television series, documentaries, cooking demonstrations, sports and fitness). Then we use TransNet V2~\cite{soucek2024transnet} to detect shot transitions and crop out massive single-shot videos. To merge the single-shot videos captured in the same scene, we employ a scene segmentation method~\cite{wu2022scene} that could understand the storyline of
the video to figure out where a scene starts and ends. It might cluster videos spanning tens of minutes within the same scene into a single group. We further sample multi-shot videos using the following strategy: shot count ranges from 1 to 5, frame count ranges from 77 to 308~(i.e., 5-20 seconds at 15 fps), with priority given to samples with higher frame counts and more shots.

\noindent\textbf{Caption Definition}: 
To provide users with the ability to define characters and customize each shot, we employ a hierarchical caption structure: global caption and per-shot captions, following~\cite{guo2025long}. We first use Gemini-2.5~\cite{comanici2025gemini} to understand the entire multi-shot video and produce a global caption, where each subject is denoted by ``Subject $\text{X}, \text{X}\in[1,2,3,\cdots]$". Subsequently, based on the global caption and each shot video, we employ Gemini-2.5 to reason the per-shot captions, using the predefined ``Subject $\text{X}$" across all captions. 
It helps the model understand cross-shot subject consistency. \textit{More details could be found in Appendix~\ref{supp:hierarchical_caption}.}

\noindent\textbf{Reference Images Collection}: We first apply YOLOv11 \cite{khanam2024yolov11}, ByteTrack~\cite{zhang2022bytetrack} and SAM~\cite{kirillov2023segment} to detect, track and segment subject images. Due to the presence of shot transitions, the tracking process is conducted shot-by-shot. This process produces the shot-level track IDs and their bounding box sequences. To merge the tracking results across all shots, we choose the largest subject image of each shot-level track ID from each shot, and group them using Gemini-2.5~(Details could be found in Appendix~\ref{supp:merge_id}). In this way, we obtain complete multi-shot tracking results and the corresponding subject images. In addition, we feed the first frames of each shot and its foreground mask into OmniEraser~\cite{wei2025omnieraser} to obtain clean backgrounds.

%% file: 4_experiment.tex
\begin{figure*}[!t]
  \centering
  \includegraphics[width=\linewidth]{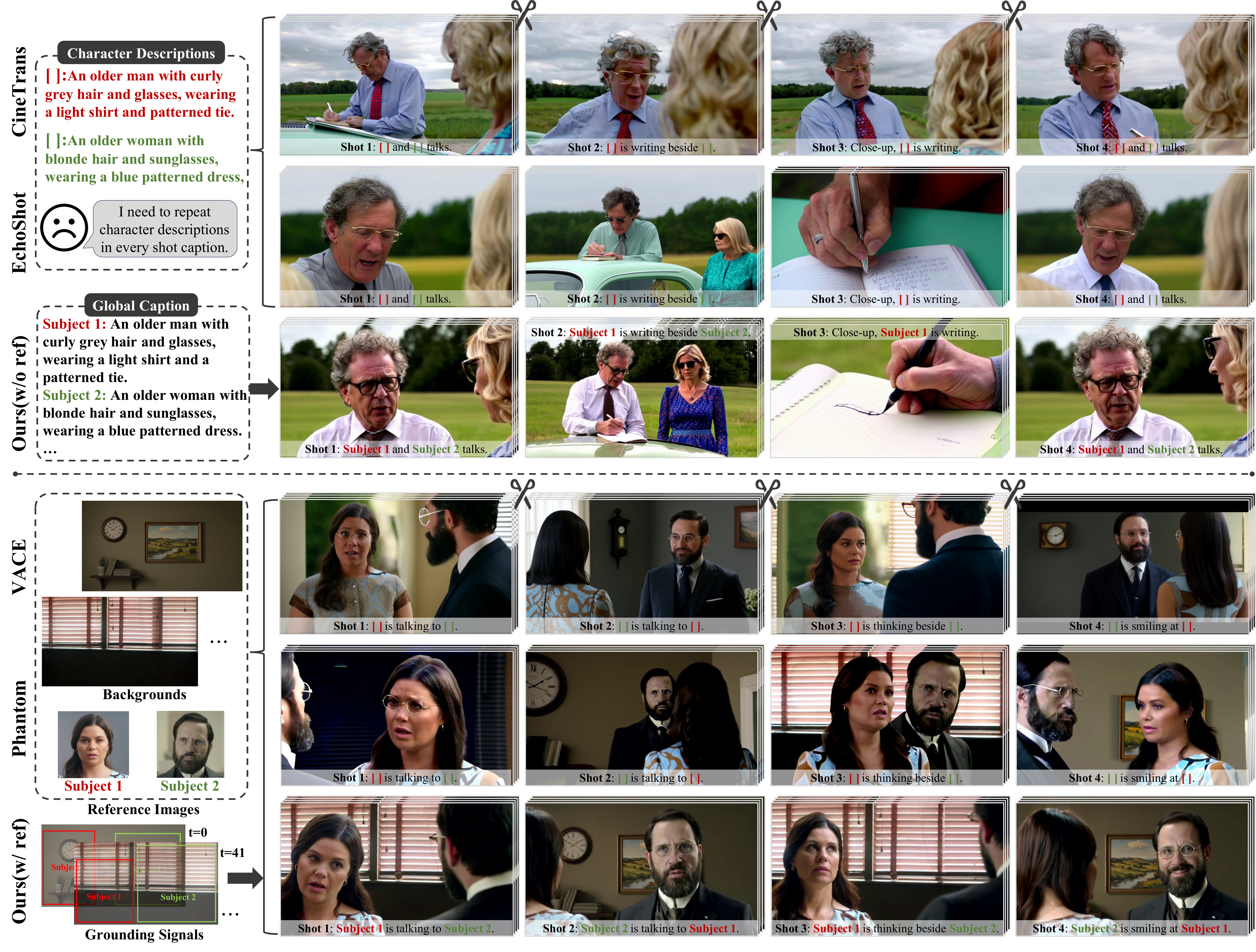}
  \vspace{-18pt} 
  \caption{\textbf{Qualitative Comparisons.} We compare with two multi-shot video generation methods~\cite{wu2025cinetrans,wang2025echoshot} in the upper part, and two single-shot reference-to-video methods~\cite{vace,liu2025phantom} under multi-shot setting in the lower part. [ ] denotes the placeholder of character descriptions for baselines. The character introductions of the bottom part are omitted for brevity.} 
  \label{fig:comparison}
  \vspace{-10pt} 
\end{figure*}

\section{Experiment}
\label{sec:experiment}
\subsection{Experimental Setup}
\noindent\textbf{Implementation Details.} Our framework is based on a pretrained single-shot T2V model with only $\sim$1B parameters at the resolution of $384\times672$. We conduct experiments for controllable multi-shot video generation on narrative videos containing 77-308 frames at 15 fps (i.e., 5-20 seconds), with each video comprising 1-5 shots. We encode each shot separately through 3D VAE~\cite{kingma2013auto}, and employ a sliding window strategy to encode and decode shots with $>77$ frames, which maintains the alignment between pixel space and latent space for multi-shot videos. We train the model on 32 GPUs, with a learning rate of $1\times10^{-5}$, batch size 1. The angular phase shift factor of Multi-Shot Narrative RoPE is set to 0.5 by default. During inference, we set the classifier-free guidance scale~\cite{ho2022classifier} as 7.5 and DDIM~\cite{ddim} steps as 50. \textit{More details can be found in the supplementary materials.}

\noindent\textbf{Baselines.} We compare our work with two multi-shot video generation methods~\cite{wu2025cinetrans,wang2025echoshot}. CineTrans~\cite{wu2025cinetrans} is the latest open-source multi-shot narrative method. EchoShot~\cite{wang2025echoshot} focuses on identity-consistent multi-shot portrait videos, rather than narrative coherence. In addition, considering that there is no controllable multi-shot method, we employ single-shot reference-to-video methods Phantom~\cite{liu2025phantom} and VACE~\cite{vace} to generate multiple single-shot videos with individual text prompts from a story for comparison. The competing baselines are all based on Wan2.1-T2V-1.3B~\cite{wan2025wan} at the resolution $480\times832$.

\noindent\textbf{Evaluation.} To comprehensively evaluate our work, we design 100 multi-shot prompts
using Gemini-2.5~\cite{comanici2025gemini}. To ensure fairness, we process the text prompts with the corresponding style for each baseline. Given that both subject consistency and scene consistency are crucial for multi-shot reference injection, we construct 90 cases encompassing three settings: subject injection, background injection, and joint injection, with 30 cases for each setting.

\textit{Metrics.} We evaluate multi-shot narrative video generation from four perspectives: (1) \textbf{Text Alignment~(TA)}: we calculate the similarity between text features and shot features extracted by ViCLIP~\cite{wang2023internvid}. (2) \textbf{Inter-Shot Consistency}: first, we calculate the holistic semantic similarity between ViCLIP shot features. Then, we apply YOLOv11~\cite{khanam2024yolov11} and SAM~\cite{kirillov2023segment} to detect and crop subjects and backgrounds from keyframes~(first, middle, and last frames), and subsequently employ DINOv2~\cite{oquab2023dinov2} to measure subject consistency and scene consistency. (3) \textbf{Transition Deviation}: we employ TransNet V2~\cite{soucek2024transnet} to detect transitions in the generated videos and calculate the frame count deviation from the ground-truth transition timestamps. (4) \textbf{Narrative Coherence}: we employ Gemini-2.5~\cite{comanici2025gemini} to evaluate the narrative logic of multi-shot videos~(\textit{Details of this metric could be found in Appendix~\ref{supp:metric_narrative}}). In addition, we evaluate the \textbf{Reference Injection Consistency} from two perspectives: (1) We detect and crop the generated subjects and backgrounds, and calculate DINO similarity with the provided references. (2) Grounding: we detect the 2D object bounding boxes and calculate the mean Intersection over Union~(mIoU) across keyframes to measure the spatiotemporal-grounded accuracy.

\subsection{Qualitative Comparison}
As shown in Fig.~\ref{fig:comparison}, we present two different feature comparisons for multi-shot text-to-video generation and multi-shot reference-to-video generation. The simplified prompts for each shot are shown in the subtitle. CineTrans~\cite{wu2025cinetrans} uses a global caption that focuses on the scene and camera transitions, and per-shot captions. Other baselines follow the individual caption manner for each shot. All these methods need to repeat the character descriptions in every shot caption, which makes it inconvenient for users. In comparison, we adopt the user-friendly hierarchical caption structure that describes subject appearance in the global caption and uses the indexed nouns in per-shot captions.

\begin{table*}[t!]
    \centering
    \caption{\textbf{Quantitative Evaluations.} \xmarkg denotes that this feature is not supported. The upper part compares multi-shot text-to-video generation, and the lower part compares multi-shot reference-to-video generation. In comparison, we achieve superior performance across all evaluation metrics, and further provide the exceptional spatiotemporal-grounded reference injection capabilities.}
    \label{table:comparison}    
    \vspace{-5.2mm}
    \setlength{\tabcolsep}{4pt}
    \begin{center}
    \begin{tabular}{l|c|ccc|c|c|ccc}  
\toprule
 & \multirow{2}{*}{Text Align.$\uparrow$}  & \multicolumn{3}{c|}{Inter-Shot Consistency$\uparrow$}  & \multirow{2}{*}{\makecell{Transition\\Deviation$\downarrow$}}      &     \multirow{2}{*}{\makecell{Narrative\\Coherence$\uparrow$}}       & \multicolumn{3}{c}{Reference Consistency$\uparrow$} \\

&  & Semantic       & Subject      & Scene      &  &  & Subject    & Background    & Grounding    \\

\midrule
CineTrans~\cite{wu2025cinetrans}      & 0.174 & 0.683 & 0.437 & 0.389 & 5.27 & 0.496 & \xmarkg       & \xmarkg       & \xmarkg       \\      
EchoShot~\cite{wang2025echoshot}       & 0.183 & 0.617 & 0.425 & 0.346 & 3.54 & 0.213 & \xmarkg       & \xmarkg       & \xmarkg       \\
\rowcolor{mygray2!70}\rule{0pt}{2.3ex}\rule[0ex]{0pt}{0pt}Ours~(w/o Ref)          & \textbf{0.196} & \textbf{0.697} & \textbf{0.491} & \textbf{0.447} & \textbf{1.72} & \textbf{0.695} & \xmarkg      & \xmarkg      & \xmarkg      \\

\midrule

VACE~\cite{vace}           & 0.201 & 0.599 & 0.468 & 0.273 & \xmarkg      & 0.325 & 0.475 & 0.361 & \xmarkg       \\
Phantom~\cite{liu2025phantom}        & 0.224 & 0.585 & 0.462 & 0.279 & \xmarkg      & 0.362 & 0.490 & 0.328 & \xmarkg       \\

\rowcolor{mygray2!70}\rule{0pt}{2.3ex}\rule[0ex]{0pt}{0pt}Ours (w/ Ref)  & \textbf{0.227} & \textbf{0.702} & \textbf{0.495} & \textbf{0.472} & \textbf{1.41} & \textbf{0.825} & \textbf{0.493} & \textbf{0.456} & \textbf{0.594}\\

\bottomrule
\end{tabular}
\end{center}

\vspace{-6mm}
\end{table*}

In the upper part, CineTrans~\cite{wu2025cinetrans} shows limited variation in camera positioning across shot clips and fails to preserve character identity consistency. This stems from that CineTrans manipulates the attention score for shot transitions, impeding the original token interactions in pretrained attention. EchoShot~\cite{wang2025echoshot} also designs RoPE-based shot transition for generating multiple portrait video clips that mainly focuses on identity consistency, ignoring other narrative details such as inconsistent clothing colors. Our method implements text-driven cross-shot subject consistency and scene consistency. Notably, the vehicle roof occupies a small area within Shot 3, yet it still maintains consistent color with the vehicle roof in shot 2.

In the lower part, we feed subject images and background images into VACE~\cite{vace} and Phantom~\cite{liu2025phantom} for multi-shot reference-to-video generation and perform inference multiple times using individual shot captions for each shot. Since all shots are generated independently, they fail to maintain inter-shot subject consistency. For instance, in the fourth row, the woman wears different clothing between Shot 1 and Shot 3. And these methods fail to fully preserve the user-provided background reference images. In contrast, we achieve satisfactory reference-driven subject consistency and scene consistency, and further support grounding signals to control the subject injection into specified regions and background injection into specified shots.

\begin{figure}[!t]
  \centering
  \includegraphics[width=\linewidth]{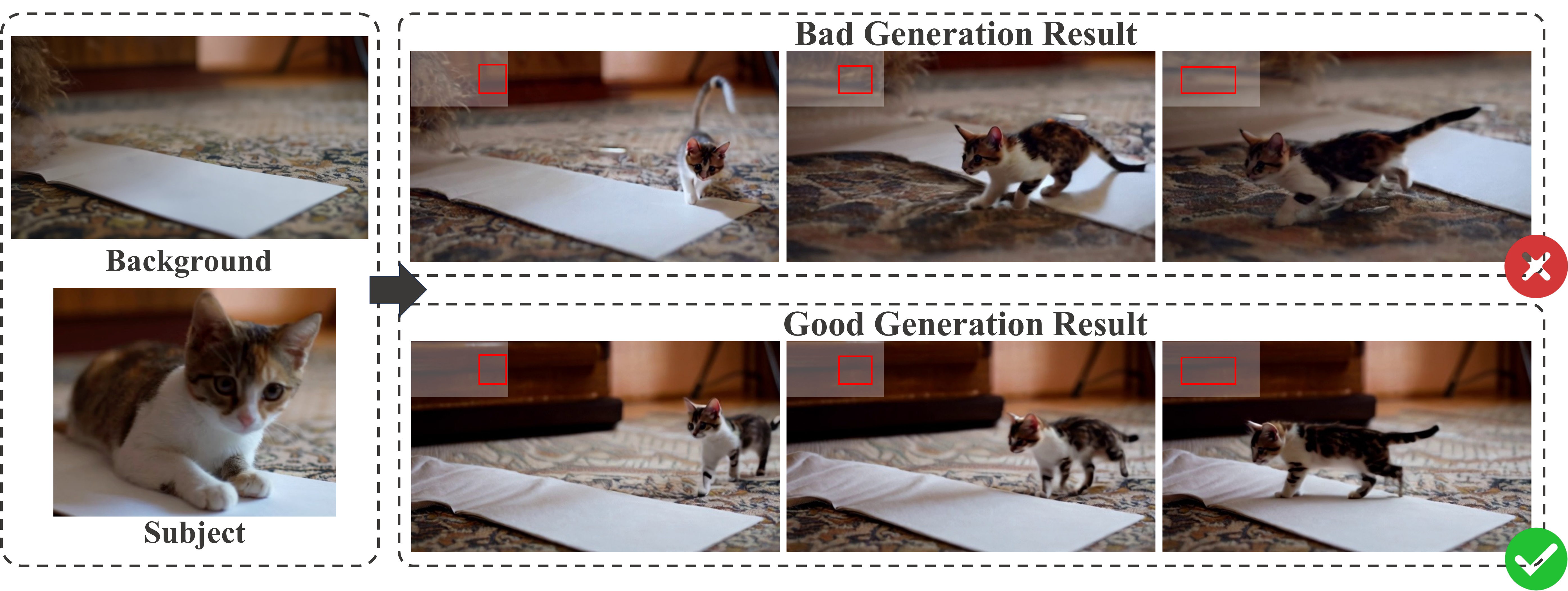}
  \vspace{-18pt} 
  \caption{Limitation visualization. We only explicitly control the subject motion, while the camera position is controlled by text prompts, which might cause the motion coupling issue.} 
  \label{fig:limit}
  \vspace{-11pt} 
\end{figure}

\subsection{Quantitative Comparison}
We report the quantitative comparison results in Table~\ref{table:comparison}. Since CineTrans~\cite{wu2025cinetrans} adds a mask matrix to the attention score, weakening the correlations across different shots, which results in unsatisfactory inter-shot consistency. On the other hand, as shown in the row 1 of Fig.~\ref{fig:comparison}, its shot transitions are not significant, leading to inferior transition deviation score and text alignment. EchoShot~\cite{wang2025echoshot} is designed for generating multiple portrait video clips rather than creating narrative content, therefore it exhibits limited narrative coherence. Benefiting from the effectiveness of the proposed framework, we achieve superior inter-shot consistency, transition deviation score and narrative coherence.

VACE~\cite{vace} and Phantom~\cite{liu2025phantom} are implemented by performing multiple independent inferences for multi-shot reference-to-video generation, so we do not calculate their transition deviation. The inter-shot consistency stems only from the text prompts and reference images, leading to suboptimal inter-shot consistency and poor narrative coherence. These methods struggle to preserve user-provided backgrounds, resulting in inferior scene consistency. In comparison, we deliver outstanding performance across all evaluation metrics while providing additional support for spatiotemporal-grounded reference injection capabilities.

\textit{Due to space constraints, we present ablation studies about key components and training strategy in  Appendix~\ref{supp:ablation}.}

%% file: 5_conclusion.tex
\section{Conclusion}
\label{sec:conclusion}
In this work, we propose MultiShotMaster, the first controllable multi-shot video generation framework. We extend a pretrained text-to-video model through two key RoPE improvements: Multi-Shot Narrative RoPE for recognizing shot boundaries and enabling controllable transitions, and Spatiotemporal Position-Aware RoPE for injecting reference tokens (subjects and backgrounds) into specific spatiotemporal regions. We also propose an automatic multi-shot \& multi-reference data curation pipeline to extract multi-shot videos, captions, cross-shot grounding signals and reference images. Our method leverages the intrinsic architectural properties to integrate text prompts, subjects, grounding signals, and backgrounds for flexible multi-shot video generation with superior controllability. We anticipate that this work could inspire future research in controllable multi-shot video generation.

\noindent\textbf{Limitation and Future Works.}
Although our method exhibits strong controllability, several key limitations require additional research to address: (1) We experiment on a pretrained single-shot T2V model with only $\sim$1B parameters at the resolution of $384\times672$, which lags behind current open-source models like the WAN family of models~\cite{wan2025wan} at the resolution $480\times832$ used by baselines. Therefore, the generation quality still needs improvement. In the future, we will implement our work on WAN 2.1/2.2 and release code. (2) We explicitly control the subject motion, while the camera position is controlled by text prompts. As shown in Fig~\ref{fig:limit}, although the generated video aligns the grounding signals, this is a consequence of the camera and object moving together. We leave this coupling issue as future work.

%% file: supp.tex
\appendix

\part*{Appendix}

\section{More Implementation Details}
\subsection{Details in Temporal Attention}
\label{supp:attnt}
To clarify the designs in temporal attention, including Multi-Shot Narrative RoPE and Spatiotemporal Position-Aware RoPE, we provide an Algorithm~\ref{al:attnt}. Specifically, the complete in-context latents $Z$ contain multi-shot video latents $z=[z_i]_{i=1}^{N_{shot}}$ and reference latents $z^{ref}=[z^m]_{m=1}^{N_{ref}}$. $N_{shot}$ represents shot count, $N_{ref}$ represents the number of input reference images (subjects and backgrounds). The input bounding box sequences of references $[boxes]_b^{N_{box}}$ contain $N_{box}$ bounding boxes. Each bounding box is represented as $[(m, t, x_1, y_1, x_2, y_2)]$, indicating the bounding box of $m$-th reference at $t$-th frame. Note that for background references, the bounding boxes are fixed as $(m, t, 0, 0, H, W)$, where $t$ is the first frame of the corresponding shot.

In temporal attention, the linear projections $to\_q$, $to\_k$, $to\_v$ first transform in-context latents $Z$ to $\tilde{Q}$, $\tilde{K}$, $\tilde{V}$. Then, by applying the Multi-Shot Narrative RoPE (i.e., Eq.~2 in the main paper), the query and key of each shot are introduced explicit shot transition signals, while keeping the narrative temporal order. For $m$-th reference containing $N_{box}^m$ boxes, we copy the query and key of the $m$-th reference $N_{box}^m$ times. Each copy is then applied with a Spatiotemporal Position-Aware RoPE based on the corresponding box in $[boxes]_b^{N_{box}^m}$. Since RoPE is not applied to value component in attention mechanism, we copy the value for attention computation. After the attention computation with the proposed multi-shot \& multi-reference attention mask, we aggregate the $N_{box}^m$ reference copies for $m$-th reference by taking their mean. Finally, the multi-shot video $\hat{z}$ and the reference latents $\bar{z}^{ref}$ are concatenated along the token dimension and fed into the linear projection $to\_out$. The attention output $Z^*$ maintains the same dimension with the input in-context latents $Z$.

\begin{algorithm}[t]
{\linespread{1.2}\selectfont
\caption{\textbf{Temporal Attention} with Multi-Shot Narrative RoPE and Spatiotemporal Position-Aware RoPE.}
\label{al:attnt}
\textbf{Input}:
\begin{itemize}
    \item In-context latents $Z$ containing:
    \begin{itemize}
        \item Multi-shot video latents $z=[z_i]_{i=1}^{N_{shot}}$
        \item Reference latents $z^{ref}=[z^m]_{m=1}^{N_{ref}}$
    \end{itemize}
    \item Bounding box sequences of references (subjects and backgrounds) $[boxes]_b^{N_{box}} = [(m, t, x_1, y_1, x_2, y_2)]_b^{N_{box}}$ 
\end{itemize}}
{\linespread{1.3}\selectfont
\textbf{Output}: In-context latents $Z^*$ after temporal attention\\
\vspace{-15pt}
\begin{algorithmic}[1] 
\STATE $\tilde{Q} = to\_q(Z)$, $\tilde{K} = to\_k(Z)$, $\tilde{V} = to\_v(Z)$ \\
\textcolor[gray]{0.4}{\textit{// Apply Multi-Shot Narrative RoPE}}
\STATE $Q = [Q_i]_{i=1}^{N_{shot}} = Eq.~2([\tilde{Q}_i]_{i=1}^{N_{shot}})$\\
$K = [K_i]_{i=1}^{N_{shot}} = Eq.~2([\tilde{K}_i]_{i=1}^{N_{shot}})$\\
$V = [\tilde{V}_i]_{i=1}^{N_{shot}}$\\
\textcolor[gray]{0.4}{\textit{// Apply Spatiotemporal Position-Aware RoPE}}
\STATE {\small $Q^{ref} = [Q_b^{ref}]_{b=0}^{N_{box}} = Eq.~3(Copy(\tilde{Q}^{ref}),[boxes]_b^{N_{box}})$} \\
{\small $K^{ref} = [K_b^{ref}]_{b=0}^{N_{box}} = Eq.~3(Copy(\tilde{K}^{ref}),[boxes]_b^{N_{box}})$} \\
{\small $V^{ref} = [\tilde{V}_b^{ref}]_{b=0}^{N_{box}} = Copy(\tilde{V}^{ref},[boxes]_b^{N_{box}})$}\\
\textcolor[gray]{0.4}{\textit{// Attention Computataion}}\\
\STATE $\hat{Z}$ = Attention($[Q, Q^{ref}], [K, K^{ref}], [V, V^{ref}]$, Mask)\\
\textcolor[gray]{0.4}{\textit{// Reference Aggregation}}\\
\STATE $\bar{z}^{ref}=[\bar{z}^m]_{m=1}^{N_{ref}}=[~mean([\hat{z}^m]_b^{N_{box}^m})~]_{m=1}^{N_{ref}}$
\STATE $Z^*=to\_out([\hat{z},\bar{z}^{ref}])$
\STATE \textbf{return} $Z^*$
\end{algorithmic}}
\end{algorithm}

\subsection{Training Paradigm}
\label{supp:train_paradigms}
The three-stage training paradigm consists of: (1) we finetune the temporal attention for spatiotemporal-specified reference injection on 300k single-shot video data with 30 epochs, batch size 8, while keeping other model parameters frozen. (2) we finetune temporal attention, cross attention and FFN on 235k multi-shot \& multi-reference data with 3 epochs, batch size 1. (3) following the second stage, we assign $(2\times)$ loss weight to subject regions and $(1\times)$ to backgrounds to train 0.5 epoch. We conduct ablation study for the training paradigm in Sec~\ref{ablation:train_paradigm} and Table~\ref{table:train_strategy}.

\subsection{Labeling Hierarchical Captions}
\label{supp:hierarchical_caption}
\begin{figure*}[!t]
  \centering
  \includegraphics[width=\linewidth]{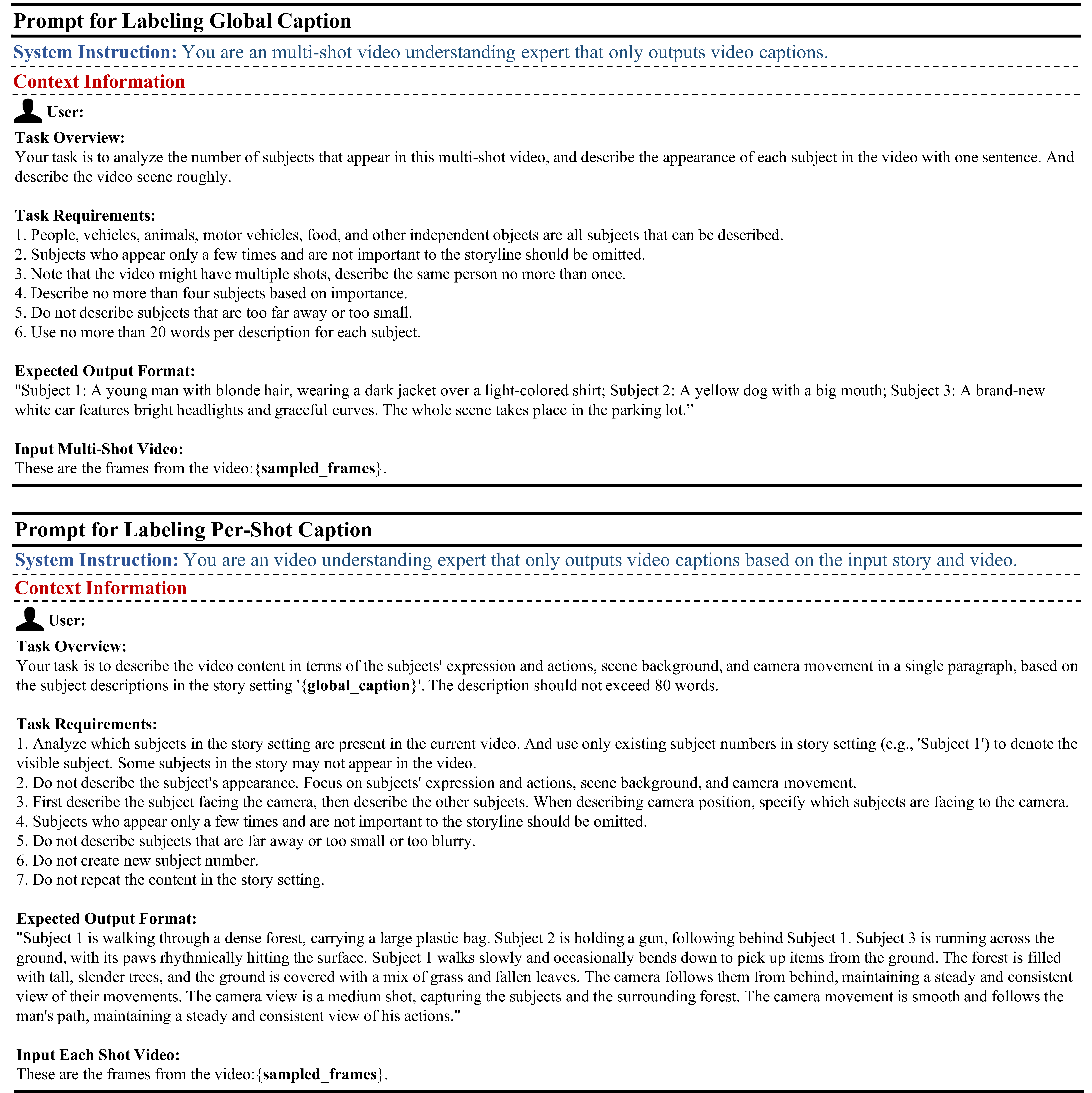}
  \vspace{-14pt} 
  \caption{Prompts of labeling global caption and per-shot captions. We first label the global caption by sampling frames from the input multi-shot video. Then we label the per-shot caption one by one.}
  \label{fig:global_local_caption}
  \vspace{-5pt} 
\end{figure*}

\begin{figure*}[!t]
  \centering
  \includegraphics[width=\linewidth]{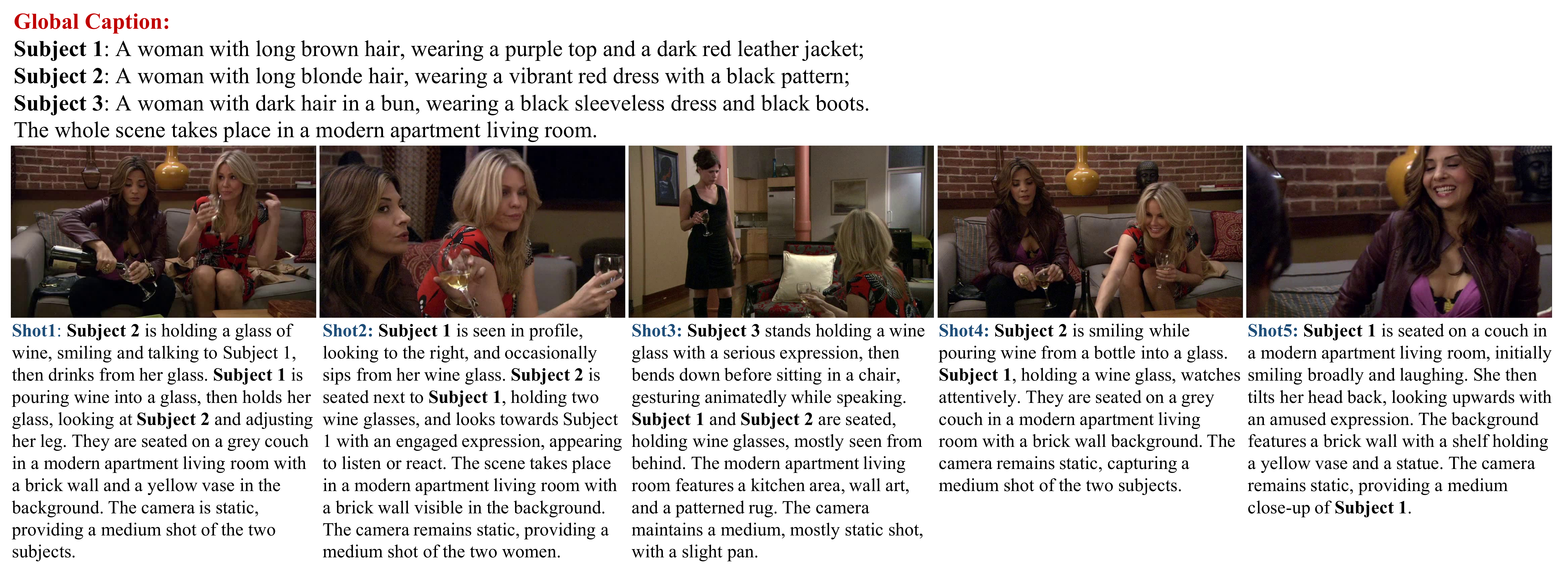}
  \vspace{-14pt} 
  \caption{Multi-shot video data example. By employing Gemini-2.5~\cite{comanici2025gemini} with the carefully-designed prompts as shown in Fig.~\ref{fig:global_local_caption}, the labeled subjects could be consistent in global and per-shot captions.}
  \label{fig:show_caption}
  \vspace{-5pt} 
\end{figure*}

As introduced in Sec~3.5 of the main paper, we employ Gemini-2.5~\cite{comanici2025gemini} to label the global caption and per-shot captions. The prompt template is shown in Fig.~\ref{fig:global_local_caption}. We begin by proportionally sampling 20 frames from the multi-shot video, ensuring at least one frame is extracted from each shot, and use Gemini-2.5 to produce a comprehensive global caption. Then we employ Gemini-2.5 to reason the per-shot captions based on the global caption and each shot video (with a sampling frame stride of 15). Each subject is denoted by ``Subject $\text{X}, \text{X}\in[1,2,3]$". As shown in Fig.~\ref{fig:show_caption}, the cross-shot consistency of subject annotations is satisfactory due to the powerful Gemini-2.5 and our carefully-designed prompt template.

\subsection{Merge Cross-Shot Tracking Annotations}
\label{supp:merge_id}
\begin{figure*}[!t]
  \centering
  \includegraphics[width=\linewidth]{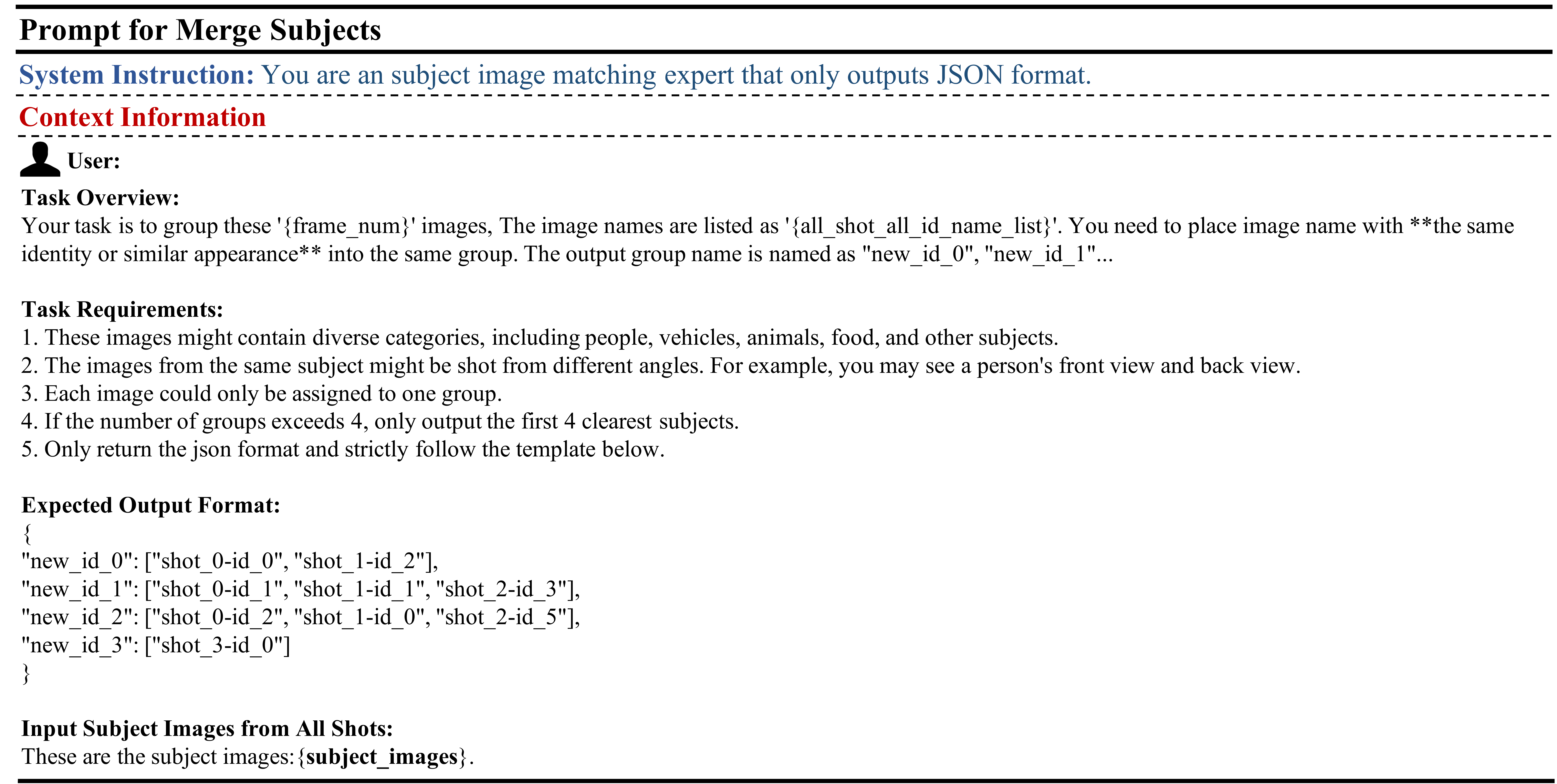}
  \vspace{-14pt} 
  \caption{By employing Gemini-2.5~\cite{comanici2025gemini} to group the subject images, we obtain complete multi-shot tracking results.}
  \label{fig:merge_id}
  \vspace{-5pt} 
\end{figure*}

As introduced in Sec~3.5 of the main paper, we conduct the tracking process shot-by-shot to obtain the bounding box sequence of each subject. To merge the cross-shot tracking results, we use Gemini-2.5~\cite{comanici2025gemini} to group the subject images by prompting with our carefully-designed prompt template as shown in Fig.~\ref{fig:merge_id}.

\begin{table}[t!]
    \centering
    \caption{\textbf{Ablation study for Multi-Shot RoPE.} We experiment on multi-shot text-to-video generation without reference input.}
    \label{table:ms_rope}    
    \setlength{\tabcolsep}{1.5pt}
    \resizebox{\linewidth}{!}{
    \begin{tabular}{l|ccc|c|c}  
\toprule
 & \multicolumn{3}{c|}{Inter-Shot Consistency$\uparrow$}  & \multirow{2}{*}{\makecell{Transition\\Deviation$\downarrow$}}      &     \multirow{2}{*}{\makecell{Narrative\\Coherence$\uparrow$}}       \\

 & Semantic       & Subject      & Scene      &  &    \\
\midrule
     
w/o MS RoPE    & \textbf{0.702} & 0.486 & \textbf{0.455} & 4.68 & 0.645     \\
\rowcolor{mygray2!70}\rule{0pt}{2.3ex}\rule[0ex]{0pt}{0pt}Ours~(w/o Ref)          & 0.697 & \textbf{0.491} & 0.447 & \textbf{1.72} & \textbf{0.695}     \\
\bottomrule
\end{tabular}}
\end{table}

\subsection{Narrative Coherence}
\label{supp:metric_narrative}
\begin{figure*}[!t]
  \centering
  \includegraphics[width=\linewidth]{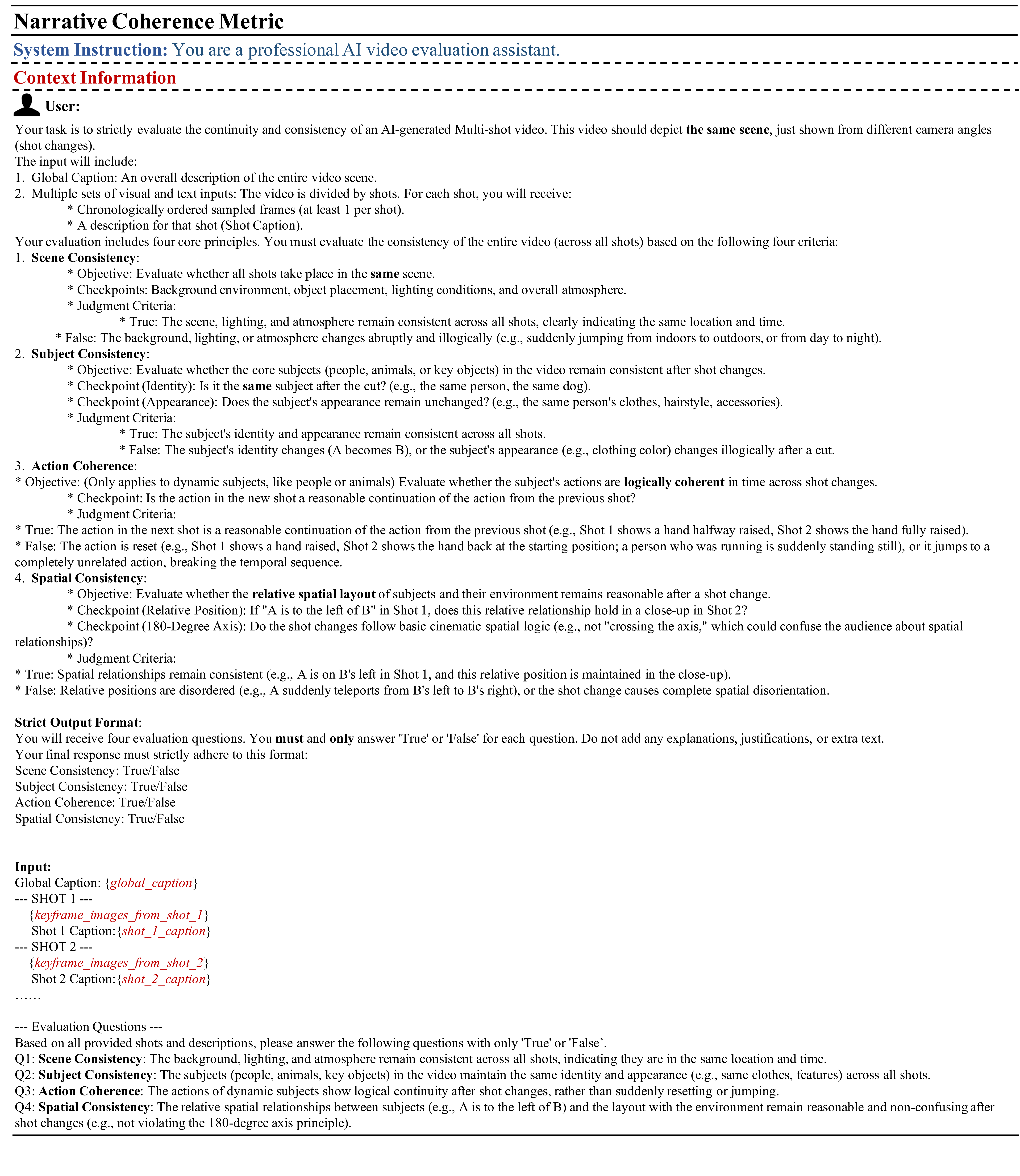}
  \vspace{-14pt} 
  \caption{We require Gemini-2.5~\cite{comanici2025gemini} to strictly adhere to cinematic narrative logic and scrutinize cross-shot content across four core dimensions: Scene Consistency, Subject Consistency, Action Coherence, and Spatial Consistency.}
  \label{fig:metric_narrative}
  \vspace{-5pt} 
\end{figure*}

To comprehensively assess the narrative coherence of generated multi-shot videos, we employ Gemini-2.5~\cite{comanici2025gemini} to construct an automated evaluation metric. We begin by proportionally sampling 20 frames from the multi-shot video, ensuring at least one frame is extracted from each shot. Subsequently, we input these frames and the hierarchical captions as a pair into Gemini-2.5. We require Gemini-2.5 to strictly adhere to cinematic narrative logic and scrutinize cross-shot content across four core dimensions: Scene Consistency, Subject Consistency, Action Coherence, and Spatial Consistency, by the constructed elaborate instructions as shown in Fig.~\ref{fig:metric_narrative}.

Specifically, Scene Consistency verifies the stability of the background, lighting, and atmosphere during transitions to ensure all shots depict the same setting; Subject Consistency strictly scrutinizes identity features and appearance attributes by comparing core objects across different viewpoints to detect unintended deviations; Action Coherence focuses on evaluating the temporal logic of dynamic behaviors to determine whether actions in subsequent shots constitute reasonable continuations of preceding ones; and Spatial Consistency examines whether the topological structure of relative positional relationships between subjects remains constant in accordance with cinematic language. Functioning as a binary classifier, the model outputs a ``True" or ``False" verdict for each dimension, thereby quantifying the generative model's capability in handling complex multi-shot spatiotemporal consistency.

\section{Ablation Study}
\label{supp:ablation}
\subsection{Ablation Study for Network Design}
We experiment with different settings to validate the effectiveness of the proposed designs in our framework:
\begin{itemize}[leftmargin=*]
\item ``w/o MS RoPE": without Multi-Shot Narrative RoPE, the shot transitions rely only on the per-shot captions.
\item ``w/o Mean": this setting randomly selects one copy from multiple copies of subject tokens after 3D attention, instead of averaging.
\item ``w/o Attn Mask": without Multi-Shot \& Multi-Reference Attention Mask, this setting uses full attention along the temporal dimension.
\item ``w/o STPA RoPE": without the Spatiotemporal Position-Aware RoPE, this setting directly concatenates the reference tokens along the temporal dimension and applies the RoPE(t=0,h,w) to each reference.
\item ``Ours~(w/o Ref)": this setting is trained using all the proposed designs, and infers multi-shot text-to-video generation without reference input.
\item ``Ours~(w/ Ref)": this setting uses the same trained checkpoint as ``Ours~(w/o Ref)" and infers multi-shot reference-to-video generation.
\end{itemize}

Since the spatiotemporal-grounded reference injection might facilitate shot transitions, we do not provide reference input to compare with ``w/o MS RoPE" setting. It relies only on the variations between per-shot captions to guide shot transitions, and uses the continuous RoPE to all frames of multi-shot videos in the temporal order. This setting cannot implement precise shot transitions by text prompts only, leading to unsatisfactory transition deviation score, as shown in Table~\ref{table:ms_rope}. Due to the lack of shot transitions, there is almost no change between shots, resulting in higher semantic and scene consistency scores. With the proposed Multi-Shot Narrative RoPE, we can perform shot transition at user-specified timestamps with superior deviation score.

We further evaluate the designs in spatiotemporal-grounded reference injection. In addition to the mentioned metrics in the main paper, we further introduce Aesthetic Score~\cite{schuhmann2022laion} to measure the aesthetic quality of the generated multi-shot videos. ``w/o Mean" might cause information loss, showing suboptimal results, as shown in Table~\ref{table:ref_inject}. the excessively long contexts in ``w/o Attn Mask" setting have unnecessary interactions between in-context tokens, leading to weak aesthetic score and reference consistency. ``w/o STPA RoPE" cannot designate the specific shot or exact spatiotemporal position where the subjects and backgrounds appear, relying only on text prompts for positioning. It shows poor performance on reference consistency. Taking advantage of the effectiveness of the proposed designs, our method shows best performance on all metrics.

\begin{table}[t!]
    \centering
    \caption{\textbf{Ablation study for reference injection.} We experiment on multi-shot reference-to-video generation.}
    \label{table:ref_inject}    
    \setlength{\tabcolsep}{1.5pt}
    \resizebox{\linewidth}{!}{
    \begin{tabular}{l|c|c|ccc}
\toprule
 & \multirow{2}{*}{\makecell{Aesthetic\\Score$\uparrow$}}    &     \multirow{2}{*}{\makecell{Narrative\\Coherence$\uparrow$}}  & \multicolumn{3}{c}{Reference Consistency$\uparrow$}       \\

& & & Subject & Scene & Grounding  \\

\midrule
     
w/o Mean    & 3.84 & 0.796 & 0.482 & 0.452 & 0.557     \\
w/o Attn Mask    & 3.72 & 0.787 & 0.468 & 0.414 & 0.561     \\
w/o STPA RoPE    & 3.79 & 0.761 & 0.425 & 0.363 & \xmarkg     \\
\rowcolor{mygray2!70}\rule{0pt}{2.3ex}\rule[0ex]{0pt}{0pt}Ours~(w/ Ref)          & \textbf{3.86} & \textbf{0.825} & \textbf{0.493} & \textbf{0.456} & \textbf{0.594}     \\

\bottomrule
\end{tabular}}

\end{table}

\begin{table*}[t!]
    \centering
    \caption{Ablation Study for training paradigm. We experiment on multi-shot reference-to-video generation. The $1^{st}/2^{nd}$ best results of settings are indicated in \underline{underline}/\textbf{bold}.
    }
    \label{table:train_strategy}    
    
    \setlength{\tabcolsep}{3pt}

    \resizebox{1\textwidth}{!}{
    \begin{tabular}{l|c|ccc|ccc}
    
\toprule
 & \multirow{2}{*}{Text Align.$\uparrow$}  & \multicolumn{3}{c|}{Inter-Shot Consistency$\uparrow$}       & \multicolumn{3}{c}{Reference Consistency$\uparrow$} \\

&  & Semantic       & Subject      & Scene  & Subject    & Background    & Grounding    \\

\midrule

I: Multi-Shot$+$Ref. Injection          & 0.211 & 0.671 & 0.464 & 0.415   & 0.454 & 0.426 & 0.477    \\
\midrule

\multirow{2}{*}{\makecell[l]{I: Multi-Shot\\II: Multi-Shot$+$Ref. Injection}}       & \multirow{2}{*}{0.219} & \multirow{2}{*}{\underline{0.695}} & \multirow{2}{*}{0.481} & \multirow{2}{*}{0.433}    & \multirow{2}{*}{0.472} & \multirow{2}{*}{0.451} & \multirow{2}{*}{0.578}      \\

 & & & & & & & \\
 
\midrule

\multirow{2}{*}{\makecell[l]{I: Ref. Injection\\II: Multi-Shot$+$Ref. Injection}}       & \multirow{2}{*}{\underline{0.222}} & \multirow{2}{*}{0.692} & \multirow{2}{*}{\underline{0.484}} & \multirow{2}{*}{\underline{0.437}}    & \multirow{2}{*}{\underline{0.485}} & \multirow{2}{*}{\underline{0.454}} & \multirow{2}{*}{\underline{0.583}}      \\

 & & & & & & & \\
\midrule
\midrule

\multirow{2}{*}{\makecell[l]{I: Ref. Injection\\II: Multi-Shot$+$Ref. Injection\\III: Multi-Shot$+$Subject-Focused Ref. Injection}} & \multirow{3}{*}{\textbf{0.227}} & \multirow{3}{*}{\textbf{0.702}} & \multirow{3}{*}{\textbf{0.495}} & \multirow{3}{*}{\textbf{0.472}}  & \multirow{3}{*}{\textbf{0.493}} & \multirow{3}{*}{\textbf{0.456}} & \multirow{3}{*}{\textbf{0.594}}\\

 & & & & & & & \\

 & & & & & & & \\

\bottomrule
\end{tabular}}    
\end{table*}

\subsection{Ablation Study for Training Paradigm}
\label{ablation:train_paradigm}
We conduct ablation study for the three-stage training paradigm introduced in Sec~3.4 of the main paper. We first explore the order of multi-shot video generation and reference-to-video generation, then shows the performance of the subject-focused post-training. The first setting involves fintuning the pretrained text-to-video generation model to learn both multi-shot task and reference-to-video task simultaneously. However, because the diffusion loss is computed across all frames to optimize global consistency, this unified training paradigm shows inadequate for effectively learning both tasks. The second setting is first learning multi-shot text-to-video generation, followed by multi-shot reference-to-video generation, both using the curated multi-shot \& multi-reference data. This setting achieves slightly lower subject consistency due to insufficient exposure to diverse subjects during training. Since the construction cost of multi-shot \& multi-reference data is relatively high, we first train the model to learn spatiotemporal-grounded reference injection task on single-shot data, and then learn both tasks using the curated multi-shot \& multi-reference data. It achieves better results on most metrics. Furthermore, we introduce subject-focused post-training that guides the model to prioritize subjects requiring higher consistency, which also promotes the modeling of cross-shot subject variations.

